\documentclass[conference]{IEEEtran}
\IEEEoverridecommandlockouts
\usepackage{graphicx} 
\usepackage[T1]{fontenc}    
\usepackage{hyperref}       
\usepackage{url}            
\usepackage{booktabs}       
\usepackage{amsfonts}       
\usepackage{nicefrac}       
\usepackage{microtype}      
\usepackage{cite}
\usepackage{amsmath,amssymb,amsfonts}
\usepackage{textcomp}
\usepackage{xcolor}
\usepackage{algorithm}
\usepackage{algpseudocode}
\usepackage{verbatim}
\usepackage{tabularx}

\def\BibTeX{{\rm B\kern-.05em{\sc i\kern-.025em b}\kern-.08em
    T\kern-.1667em\lower.7ex\hbox{E}\kern-.125emX}}
\algdef{SE}[SUBALG]{Indent}{EndIndent}{}{\algorithmicend\ }%
\algtext*{Indent}
\algtext*{EndIndent}
\begin{document}

\title{Poisson Flow Consistency Training}

\author{\IEEEauthorblockN{
    Anthony Zhang\IEEEauthorrefmark{1},
    Mahmut Gokmen\IEEEauthorrefmark{2},
    Dennis Hein\IEEEauthorrefmark{3},
    Rongjun Ge\IEEEauthorrefmark{4},
    Wenjun Xia\IEEEauthorrefmark{4},
    Ge Wang\IEEEauthorrefmark{4},
    and Jin Chen\IEEEauthorrefmark{5}
    }
\and
\IEEEauthorblockA{\IEEEauthorrefmark{1}Pratt School of Engineering \\ Duke University \\ Durham, North Carolina, United States} \\
\IEEEauthorblockA{\IEEEauthorrefmark{4}Biomedical Imaging Center \\ Rensselaer Polytechnic Institute \\ Troy, New York, United States}
\and
\IEEEauthorblockA{\IEEEauthorrefmark{2}Department of Computer Science \\ University of Kentucky \\ Lexington, Kentucky, United States} \\
\IEEEauthorblockA{\IEEEauthorrefmark{5}Department of Medicine \\ Department of Biomedical Informations \\ University of Alabama at Birmingham \\ Birmingham, Alabama, United States}
\and
\IEEEauthorblockA{\IEEEauthorrefmark{3}School of Engineering Sciences \\ Department of Physics \\ KTH Royal Institute of Technology \\ Stockholm, Sweden} 
}

\maketitle
\begin{abstract}
    The Poisson Flow Consistency Model (PFCM) is a consistency-style model based on the robust Poisson Flow Generative Model++ (PFGM++) which has achieved success in unconditional image generation and CT image denoising. Yet the PFCM can only be trained in distillation which limits the potential of the PFCM in many data modalities. The objective of this research was to create a method to train the PFCM in isolation called Poisson Flow Consistency Training (PFCT). The perturbation kernel was leveraged to remove the pretrained PFGM++, and the sinusoidal discretization schedule and Beta noise distribution were introduced in order to facilitate adaptability and improve sample quality. The model was applied to the task of low dose computed tomography image denoising and improved the low dose image in terms of LPIPS and SSIM. It also displayed similar denoising effectiveness as models like the Consistency Model. PFCT is established as a valid method of training the PFCM from its effectiveness in denoising CT images, showing potential with competitive results to other generative models. Further study is needed in the precise optimization of PFCT and in its applicability to other generative modeling tasks. The framework of PFCT creates more flexibility for the ways in which a PFCM can be created and can be applied to the field of generative modeling.
\end{abstract}
\section{Introduction}

Deep Generative Modeling leverages deep neural networks to synthesize highly realistic data, achieving notable success in audio \cite{kong_hifi-gan_2020, pascual_full-band_2023} and image \cite{song_consistency_2023, goodfellow_generative_2014} generation. Diffusion Models \cite{ho_denoising_2020, song_score-based_2021, karras_elucidating_2022} have significantly advanced the field; in particular, physically inspired diffusion-style models have set new benchmarks in generative AI. A key innovation, the Poisson Flow Generative Model++ (PFGM++) \cite{xu_pfgm_2023}, inspired by high-dimensional electrostatics, unified Diffusion Models \cite{song_score-based_2021, karras_elucidating_2022} with the Poisson Flow Generative Model (PFGM) \cite{xu_poisson_2022} and achieved the state-of-the-art in both unconditional and class-conditional image generation, as well as permitting the tuning between robustness and rigidity of the model.

To enable fast sampling and reduced computational demands, models have been developed that can sample images with a single input to model, while retaining the advantages of Diffusion Models. The Consistency Model (CM) \cite{song_consistency_2023}, inspired by Diffusion Models \cite{song_score-based_2021}, has emerged as a competitive alternative in image generation. CMs can be trained either through distillation, which requires a pre-trained Diffusion Model, or in isolation, which does not rely on a pre-trained model. Similarly, the Poisson Flow Consistency Model (PFCM) \cite{hein_pfcm_2025} has been introduced to sample in a single step with the advantages of PFGM++. It has demonstrated success in computed tomography (CT) image denoising, achieving competitive results with PFGM++ and other Diffusion Models while sampling in only a single-step. 

However, the training process of PFCMs is currently limited to the distillation approach, necessitating a pre-trained PFGM++. As such, if PFGM++ is unavailable or prohibitively expensive to train, PFCM cannot be trained. To date, no method has been developed to train a PFCM in isolation.

This paper introduces the first practical methodology for training PFCM in isolation. To enhance model stability and optimize training performance, we incorporate techniques from Improved Consistency Training (iCT) \cite{song_improved_2023} along with a set of advanced conditional optimization strategies, including a sinusoidal discretization schedule and a Beta noise distribution. 
The proposed isolated training approach is validated using the Low-Dose CT Image Denoising and Projection Dataset \cite{moen_low-dose_2020}, demonstrating its effectiveness in achieving robust performance without relying on pre-trained models.

In this paper, we make the following contributions: 1) We introduce a stable and effective method for training the PFCM in isolation which we name Poisson Flow Consistency Training (PFCT); 2) We evaluate the effectiveness of PFCT for low-dose computed tomography (LDCT) denoising.

\section{Background}
\subsection{Low Dose Computed Tomography Denoising}

Computed tomography (CT) is one of the most prominent medical imaging modalities. It offers the ability to scan patients in even a few seconds, and is key in medical diagnoses. CT imaging requires the use of x-rays, causing CT scans to become an increasingly notable source of ionizing radiation \cite{brenner_computed_2007, martin_health_2006}, potentially causing health issues, including cancer \cite{miglioretti_use_2013, lin_radiation_2010}, so the guidelines of As Low As Reasonably Achievable (ALARA) for radiation exposure are followed. LDCT scans can mitigate this issue, using a reduced level of radiation dose. As LDCT scans result in increased noise in reconstructed images \cite{goldman_principles_2007}, these images have worse levels of image quality, which hinders their use by radiologists. Although methods like Iterative Reconstruction \cite{silva_innovations_2010}, Wavelet Transform Domain Filtering, Bilateral Filtering, Non-Local Means Filtering, and others  \cite{diwakar_review_2018} have been introduced to resolve this issue, Deep Generative Modeling methods like the Generative Adversarial Network (GAN) \cite{gajera_ct-scan_2021, huang_du-gan_2022}, the Variational Autoencoder (VAE) \cite{fan_quadratic_2020}, and the Score-Based Generative Model \cite{song_solving_2022, wu_wavelet-improved_2024} have been used to improve the image quality of CT images. PFCMs has been particularly successful in this domain, which in distillation has previously achieved SSIM and PSNR as high as 0.98 and 45.3, respectively \cite{hein_pfcm_2025}.

\subsection{Poisson Flow Consistency Model}

The PFCM \cite{hein_pfcm_2025} applies the single-step generation capabilities of the Consistency Model \cite{song_consistency_2023} to the Poisson Flow Generative Model++ \cite{xu_pfgm_2023} to allow tuning of the hyperparameter $D$ for robustness and rigidity while only requiring one function evaluation. Treating the $N$-dimensional data as charges in $N+D$ dimensions, the PFCM uses the perturbation kernel from PFGM++, $p_{r}(\cdot|x)$, to sample noisy data. The perturbation kernel is split into a uniform angle component $\mathcal{U}_{\psi}(\psi)$ and a perturbed radius component $p_{r}(R)$. $\mathcal{U}_{\psi}$ in this case represents the uniform distribution over the angle component and $p_{r}(R)$ is the distribution of the perturbed radius $R=||x_{\sigma}-x||$ where

\begin{equation}
    p_{r}(R) \propto \frac{R^{N-1}}{(R^{2}+r^{2})^\frac{N+D}{2}}
\end{equation}
from \cite{xu_pfgm_2023}. The PFCM uses the same hyperparameter transfer as the PFGM++ for $r$, where $r=\sigma\sqrt{D}$. The noise level, $\sigma$, is found by discretizing $[\sigma_{min}, \sigma_{max}]$ into $N-1$ intervals with $\sigma_{min}=\sigma_{1} < \sigma_{2} < ... < \sigma_{N} = \sigma_{max}$. Training the PFCM with distillation involves sampling $x_{\sigma_{i+1}}$ from the perturbation kernel, and then obtaining $\Breve{x}_{\sigma_{i}}$ using a single reverse step of the probability flow ordinary differential equation (PF ODE) \cite{song_score-based_2021} with a pretrained PFGM++ model, with
\begin{equation}
    \Breve{x}_{\sigma_{i}}=x_{\sigma_{i+1}} + (\sigma_{i}-\sigma_{i+1})\Phi(x_{\sigma_{i+1}}, \sigma_{i+1}, \phi),
\end{equation}

The PFCM is trained with $\Phi_{PFGM++}(x, \sigma, \phi) := s_{\phi}(\Tilde{x}) = \sqrt{D}E(\Tilde{x}_{\sigma})_{x_{\sigma}} \cdot E(\Tilde{x}_{\sigma})_{r}^{-1}$, using the definition from \cite{xu_pfgm_2023} of the high-dimensional electric field $E(\Tilde{x})$. It uses the consistency distillation loss \cite{song_consistency_2023}, 
\begin{equation}
    \mathcal{L}(\theta, \theta^{-}; \phi) = \lambda(\sigma_{i})d(f_{\theta}(x_{\sigma_{i+1}}, \sigma_{i+1}, y), f_{\theta^{-}}(\Breve{x}_{\sigma_{i}}, \sigma_{i}, y))
\end{equation}\label{eq:consistency_distillation_loss}
where $\lambda(\cdot)$ represents a weighting function, $d(\cdot, \cdot)$ is a distance metric, and $\theta^{-}$ is a running average over past values of $\theta$.

\subsection{Improved Techniques for Training Consistency Models}

To improve the training of Consistency Models in isolation, \cite{song_improved_2023} adapted the training process to significantly increase the generation capabilities of the iCT \cite{song_improved_2023}. To assign greater weight to smaller noise levels, the weighting function is changed from $\lambda(\sigma)=1$ to $\lambda(\sigma_{i})=\frac{1}{\sigma_{i+1}-\sigma_{i}}$. For further improvement, iCT removes the exponential moving average of the teacher model to greatly improve the quality of CT across distance metrics.

While the original CM used either the squared $\ell_{2}$ metrics or LPIPS \cite{zhang_unreasonable_2018}, iCT introduced the Pseudo-Huber distance metric, 
\begin{equation}\label{eqn:pseudo-huber_loss}
    d(x, y)=\sqrt{||x-y||_{2}^{2}+c^{2}}-c
\end{equation}
where $c=0.00054\sqrt{d}$.

In order to balance between bias and variance, iCT uses an exponential discretization schedule for the noise levels, where
\begin{equation}
    N(k) = \text{min}(s_{0}2^{\frac{k}{K^{'}}}, s_{1})+1,  K^{'} = \lfloor\frac{K}{\text{log}_{2}\lfloor s_{1}/s_{0}\rfloor+1}\rfloor
\end{equation}\label{eq:exponential_schedule}
and $s_{0} = 10$ and $s_{1}=1280$. Aligning with the argument that greater weight should be given to smaller noise levels, iCT also introduces the lognormal noise distribution, 
\begin{equation}
    p(\sigma_{i})\propto \text{erf}(\frac{\text{log}(\sigma_{i+1})-P_{mean}}{\sqrt{2}P_{std}})-\text{erf}(\frac{\text{log}(\sigma_{i})-P_{mean}}{\sqrt{2}P_{std}})
\end{equation}
where $P_{mean}=-1.1$ and $P_{std} = 2.0$. With these adaptations, the iCT can achieve a Fr\'echet Inception Distance (FID) of 2.51 on CIFAR-10, and in two-step generation the model can further improve to a FID score of 2.24, surpassing diffusion distillation methods \cite{ho_denoising_2020}.

\section{Methodology}
\subsection{PFCM Training in Isolation}


To train any form of a Consistency Model, adjacent noisy data must be sampled for the loss function \eqref{eq:consistency_distillation_loss}. The PFGM++ in PFCM is used to estimate $\Breve{x}_{\sigma_{i}}$ from ${x}_{\sigma_{i+1}}$. To train in isolation, the PFGM++ must be removed from the model training.

In physical terms, the perturbation kernel is split into a uniform angle component and a perturbed radius component. Thus, we propose that two adjacent perturbed data can be sampled using only a different radius by holding the uniform angle constant for both $\Breve{x}_{\sigma_{i}}$ and $x_{\sigma_{i+1}}$. Using this method, adjacent perturbed data can be generated by sampling a single uniform angle $v = u/||u||_{2}, u \sim \mathcal{N}(0; I)$ as in equations \eqref{eqn:sample_x_i+1} and \eqref{eqn:sample_x_i}.
\begin{equation}\label{eqn:sample_x_i+1}
    x_{\sigma_{i+1}} = x + vp_{r_{i+1}}(R)
\end{equation} 
\begin{equation}\label{eqn:sample_x_i}
    x_{\sigma_{i}} = x + vp_{r_{i}}(R)
\end{equation}

Since the idea of self-consistency trained in isolation is present in both this model and iCT, we use optimizations introduced in \cite{song_improved_2023} to improve PFCM training in isolation. The exponential moving average (EMA) is removed from model training and the Pseudo-Huber Loss in \eqref{eqn:pseudo-huber_loss} is used as a distance metric. The weighting function, $\lambda(\sigma_{i}) = \frac{1}{\sigma_{i+1}-\sigma_{i}}$, is also used; the value $c=0.00054\sqrt{N}$ is kept the same, where $N$ is the data dimension.

As in \cite{song_consistency_2023}, the noise levels $[\sigma_{min}, \sigma_{max}]$ are discretized into $\sigma_{min}=\sigma_{1}<\sigma_{2}<...<\sigma_{N}=\sigma_{max}$. A sinusoidal discretization schedule, smoother than the exponential, is introduced as the timestep schedule $M(\cdot)$ \cite{gokmen_high_2024, gokmen_enhancing_2024}. The exponential discretization schedule \eqref{eq:exponential_schedule} can have large jumps in the number of timesteps as training steps increase. A sinusoidal schedule increases the schedule in a smoother manner. These smoother timestep schedules can be used to facilitate adaptability, and this paper uses a slightly adapted version of the sinusoidal timestep schedule. With $s_{0}$ as the lower bound, $s_{1}$ as the upper bound, and K as the total number of training steps, the discretization schedule $M(k)$ is defined as 

\begin{equation}\label{eqn:discretization_schedule}
    M(k) = \min(\bigg|s_{1}\sin{\frac{\left\lfloor3k\pi/K\right\rfloor}{6}} + s_{0}\bigg|+1, s_{1}+1)
\end{equation}

Following the idea that high noise scheduling is a must \cite{gokmen_high_2024}, a Beta noise distribution \cite{gokmen_enhancing_2024} is introduced to select noise levels. Given the definition of the noise boundaries from \cite{karras_elucidating_2022}, $\sigma_{i} = (\sigma_{min}^{1/\rho} + \frac{i+1}{N-1}(\sigma_{min}^{1/\rho}-\sigma_{max}^{1/\rho}))^{\rho}$, instead of using a uniform distribution to sample i, the Beta distribution is used. With a batch size $|B|$, $|B|$ samples are taken from the Beta distribution in set B, $\{b_{1}, b_{2}, ... b_{|B|}\}, b_{j} \sim \text{Beta}(\alpha, \beta)$. Then, the timestep $i$ is selected through:

\begin{equation}
    i =\left\lfloor\frac{b_{j}-\text{min}(B)}{\text{max}(B)-\text{min}(B)} * (M-1)\right\rfloor, \forall b_{j} \in B
\end{equation}

These timesteps are then used to generate the $\sigma$s for the perturbation kernel and for the hyperparameter transfer.

Using these adaptations, the algorithm for PFCT is presented below. The condition, y, is fed into the model as an additional input to better enforce learning of the prior distribution of CT images. Although it would be technically possible to learn the same distribution without a conditional image generation, this will facilitate the learning of the conditional prior distribution since the task is a conditional inverse problem.

\begin{algorithm}[H]
    \caption{Poisson Flow Consistency Training. Extended from \cite{song_consistency_2023} and \cite{hein_pfcm_2025}}\label{alg:pfct}
    \begin{algorithmic}
        \Require Dataset $\mathcal{D}$, initial model parameter $\theta$, learning rate $\eta$, step schedule $M(\cdot)$, $d(\cdot, \cdot)$, and $\lambda(\cdot)$, batch size |B|
        \Repeat
            \State Sample $(x, y) \sim \mathcal{D}$
            \State Sample $B = \{b_{1}, b_{2}, ... b_{|B|}\}, b_k \sim \text{Beta}(\alpha, \beta)$ 
            \State Set $i\sim\left\lfloor\frac{b_{k}-\text{min}(B)}{\text{max}(B)-\text{min}(B)} * (M(k)-1)\right\rfloor$
            \State Set $r_{i+1} \gets \sigma_{i+1}\sqrt{D}$ and $r_{i} \gets \sigma_{i}\sqrt{D}$
            \State Sample $R_{i+1} \sim p_{r_{i+1}}(R)$ and $R_{i} \sim p_{r_{i}}(R)$
            \State Sample uniform angles $v \gets u/||u||_{2}$, $u \sim \mathcal{N}(0; I)$
            \State $\mathcal{L}(\theta) \gets $
            \State \hspace*{6.5mm}$\lambda(\sigma_{i})d(f_{\theta}(x + vR_{i+1}, \sigma_{i+1}, y), f_{\theta}(x + vR_{i}, \sigma_{i}, y))$
            \State $\theta \gets \theta - \eta\nabla_{\theta}\mathcal{L}(\theta)$
        \Until convergence
    \end{algorithmic}
\end{algorithm}

\subsection{Implementation Details}

The model was implemented using a U-Net Style Architecture with an additional Weighted Attention Gate for improved conditional image generation \cite{gokmen_high_2024, oktay_attention_2018}. The models were trained with a learning rate of $1 * 10^{-4}$ and the RAdam optimizer \cite{liu_variance_2021}, and D=2048. $s_{0}$ was set to 10 and $s_{1}$ to 100, while $\alpha$ was set to 1.5 and $\beta$ to 5.0.

\section{Results}

\subsection{Experiments}
The dataset used for training, validation, and testing was the Low-Dose CT Image Denoising and Projection Dataset \cite{moen_low-dose_2020, mccollough_low_2020}. Abdominal CT images with a full LDCT/Full-Dose CT data pair were included in the data. These images were split into a training (11,646 images), validation (1,752 images), and testing (1,052 images) set and were normalized from their CT numbers for processing. The CT images for training used a random crop of 128x128, while the images for validation used a central crop of 128x128. The testing images used the full 512x512 image. 

The results were evaluated using the metrics Structural Similarity Index (SSIM), Peak Signal Noise Ratio (PSNR), and Learned Perceptual Image Patch Similarity (LPIPS) \cite{zhang_unreasonable_2018}. These metrics were calculated in comparison to the full dose CT image.

\subsection{Results}

\begin{figure}[H]
    \centering
    \includegraphics[width=1\linewidth]{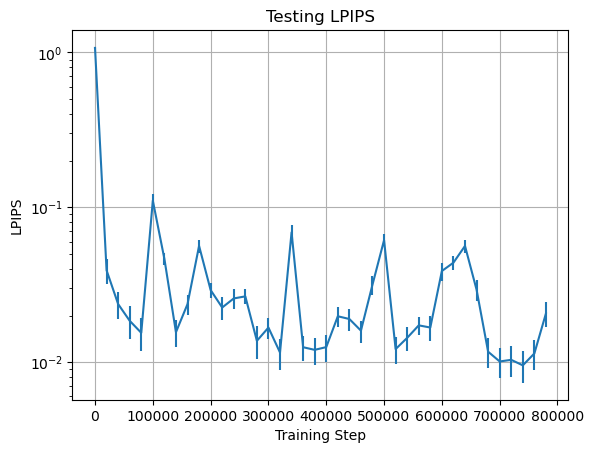}
    \caption{Testing LPIPS (logarithmic scale) over training steps.}
    \label{fig:test-results}
\end{figure}

\begin{figure}[H]
    \centering
    \includegraphics[width=1\linewidth]{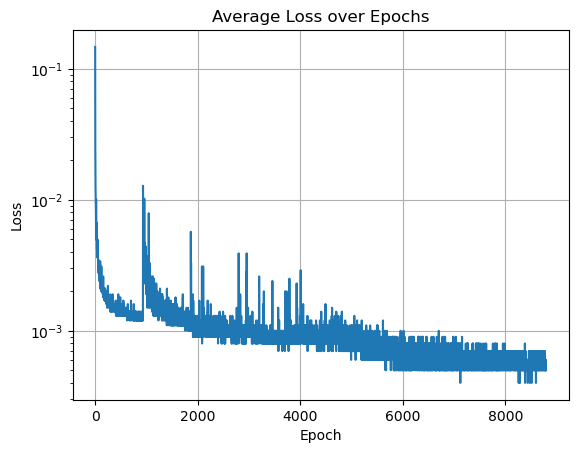}
    \caption{Average training loss (logarithmic scale) over epochs.}
    \label{fig:train-results}
\end{figure}

The training loss and testing LPIPS over the training process are displayed in Figures \ref{fig:test-results} and \ref{fig:train-results}. The training loss indicated the consistency of the outputs over epochs, while the testing LPIPS indicated the quality of the denoised images over training steps. There were visible fluctuations in both, but as training steps (or epochs) increased, the average loss and testing LPIPS gradually decreased.

Table \ref{tab:reliable_results} displays the LPIPS, SSIM, and PSNR scores of the LDCT scans, PFCT, and other published generative models. PFCT achieved a lower LPIPS compared to LDCT and a higher SSIM and PSNR. The LDCT data is from \cite{hein_pfcm_2025, gokmen_enhancing_2024} while the EDM, PFGM++, CM, ICCM, PFCM, and PFCT results are from the same dataset as \cite{hein_pfcm_2025}.

\renewcommand{\arraystretch}{1.25}

\begin{center}
\begin{table}[H]
\centering
\caption{LDCT Denoising Results}\label{tab:reliable_results}
\begin{tabularx}{.5\textwidth}{ 
    >{\centering\arraybackslash}X 
    >{\centering\arraybackslash}X
    >{\centering\arraybackslash}X
    >{\centering\arraybackslash}X
    >{\centering\arraybackslash}X
}
     \toprule
     Type & LPIPS $(\downarrow)$ & SSIM $(\uparrow)$ & PSNR $(\uparrow)$ & NFE \\ \hline
     LDCT &          
                     $0.075\pm0.022$ & $0.94\pm0.02$ & $41.50\pm1.60$ &  - \\
     \hline
     EDM \cite{karras_elucidating_2022}&           
                     $0.011\pm0.005$ & $0.97\pm0.01$ & $44.22\pm1.43$ & $79$ \\
     PFGM++ \cite{xu_pfgm_2023}&        
                     $0.011\pm0.005$ & $0.97\pm0.01$ & $44.14\pm1.47$ & $79$ \\
     \hline
     CM \cite{song_consistency_2023}& 
                     $0.016\pm0.006$ & $0.94\pm0.01$ & $41.92\pm0.99$ & $1$ \\
     ICCM \cite{gokmen_enhancing_2024}& 
                     $0.018\pm0.010$ & $0.97\pm0.01$ & $45.62\pm1.23$ & $1$ \\
     PFCM (D=128) \cite{hein_pfcm_2025}&
                     $0.009\pm0.004$ & $0.98\pm0.01$ & $45.76\pm1.53$ & $1$ \\
     \hline
     PFCT (D=2048)&
                     $0.018\pm0.005$ & $0.96\pm0.02$ & $42.07\pm2.01$ & $1$\\
     \bottomrule
     \multicolumn{5}{l}{Extended from \cite{hein_pfcm_2025, gokmen_high_2024}}

\end{tabularx}
\end{table}
\end{center}

\begin{figure}[H]
    \centering
    \includegraphics[width=1\linewidth]{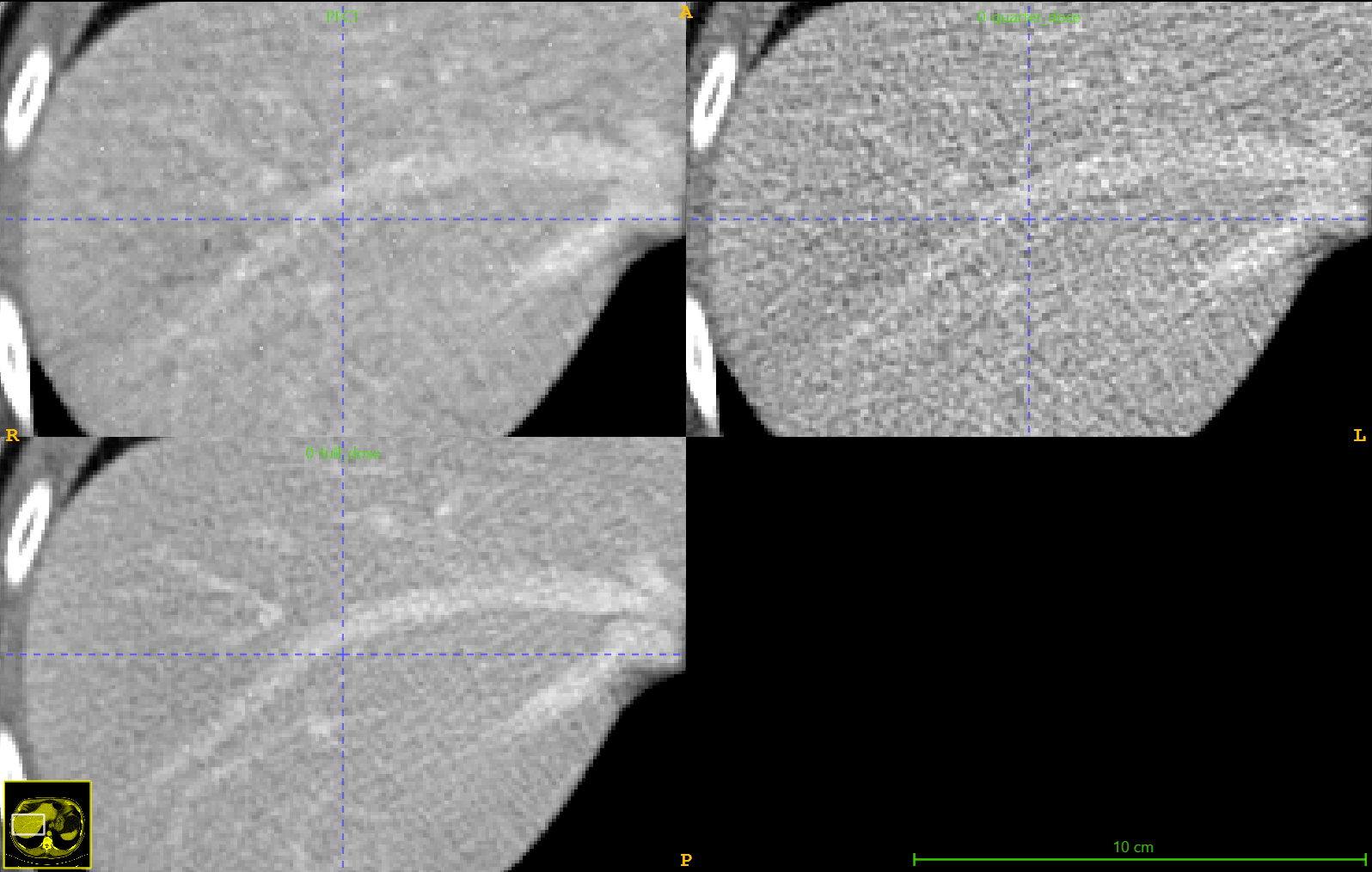}
    \caption{Top-left: PFCT, Top-right: LDCT, Bottom-left: NDCT. 3.50 pixel/mm zoom, Window: 350, Level: 50. Comparing PFCT to NDCT: SSIM .9899, PSNR 48.122; Comparing LDCT to NDCT: SSIM .9829, PSNR 46.906}
    \label{fig:qual-results}
\end{figure}

Figure \ref{fig:qual-results} displays the qualitative results of the LDCT denoising. There are texture differences between the three images present (LDCT, NDCT, and PFCT), and features like blood vessels can be seen in the same locations in each image.

\section{Discussion}

A practical method of training PFCM in isolation, PFCT, has been presented. The improvement in quantitative metrics indicated that PFCT improves the perceptual quality of LDCT images, as LPIPS is an image quality metric compatible with human perception \cite{zhang_unreasonable_2018}. PFCT caused an increase in PSNR in the generated images. As PSNR is a metric based on the mean-square error which does not necessarily correlate to human perception, an increase in PSNR does not definitively indicate a large improvement to image quality, but it did indicate that pixel-wise similarity increases \cite{zhang_unreasonable_2018}. The increase in SSIM, another standard image quality metric in CT image denoising and the decrease of LPIPS did show that PFCM denoised images as intended and increased their human perceptual quality. 

The PFCT training method did not perform as well as other generative AI methods. Both diffusion-based methods, EDM \cite{karras_elucidating_2022} and PFGM++ \cite{xu_pfgm_2023}, outperformed PFCT in LPIPS, SSIM, and PSNR, and the PFCM distillation training method had the best overall performance. However, the PFCT showed promise as a competitive training method. Since it was able to achieve similar LPIPS with methods such as ICCM \cite{gokmen_enhancing_2024} and outperform the Consistency Model \cite{song_consistency_2023} in SSIM and PSNR, the PFCT is able to be established as a method capable for medical image denoising.

With PFCT, the PFCM can function as a standalone Consistency Model, with the possibility of being trained in both ways originally described for CM: in isolation and in distillation \cite{song_consistency_2023}. Here, only LDCT denoising was investigated, yet the use of PFCT for unconditional image generation has not yet been explored. Further studies can focus on PFCT's capabilities in other data modalities.

This model is not yet in its most optimal state. With only one value of D being tested, it is very probable that the optimal value of D may further improve performance. Ablation studies are also required to further determine which optimizations have the greatest effect in improving the model.

\section{Conclusion}

Poisson Flow Consistency Training, PFCT, has been introduced as a method to train the Poisson Flow Consistency Model. There is now a method to train the PFCM without requiring a pre-trained PFGM++, allowing greater flexibility. PFCT can create a PFCM capable of denoising LDCT image denoising and is applicable to the medical imaging domain. This new training method can open possibilities for future research utilizing the PFCM.

\bibliographystyle{IEEEtran}
\bibliography{biblio}
\end{document}